# Human-Machine Interface for Remote Training of Robot Tasks.

Jordi Spranger*, Roxana Buzatoiu*, Athanasios Polydoros†, Lazaros Nalpantidis* and Evangelos Boukas*

*Robotics, Vision and Machine Intelligence Lab.,
Materials and Production Dept.,
Aalborg University Copenhagen, Denmark
Email: jordispranger@aol.com, buzatoiu.roxana@gmail.com, lanalpa@mp.aau.dk, eb@mp.aau.dk

† Intelligent and Interactive Systems Lab.,
University of Innsbruck, Innsbruck, Austria
Email:athanasios.polydoros@uibk.ac.at

*Abstract*—Regardless of their industrial or research application, the streamlining of robot operations is limited by the proximity of experienced users to the actual hardware. Be it massive open online robotics courses, crowd-sourcing of robot task training, or remote research on massive robot farms for machine learning, the need to create an apt remote Human-Machine Interface is quite prevalent. The paper at hand proposes a novel solution to the programming/training of remote robots employing an intuitive and accurate user-interface which offers all the benefits of working with real robots without imposing delays and inefficiency. The system includes: a vision-based 3D hand detection and gesture recognition subsystem, a simulated digital twin of a robot as visual feedback, and the "remote" robot learning/executing trajectories using dynamic motion primitives. Our results indicate that the system is a promising solution to the problem of remote training of robot tasks.

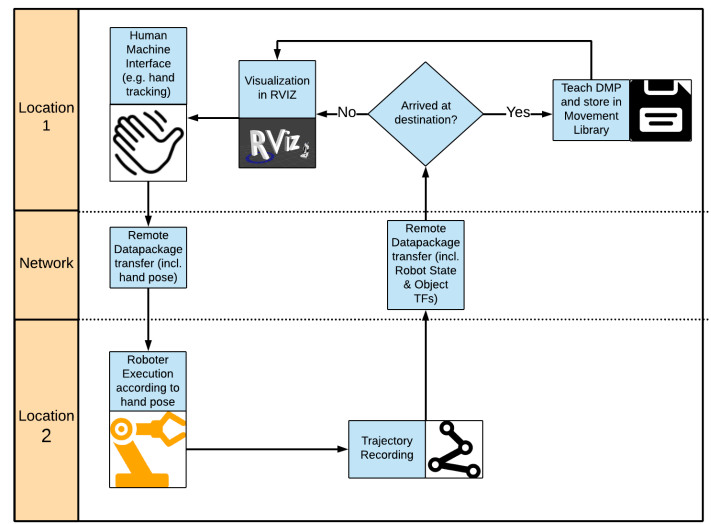

Fig. 1. The flowchart of the Human Machine Interface

## I. INTRODUCTION

Servitization concerns selling time-to-use instead of selling ownership of physical capital required for a service. A common example is the public transport where one buys a ticket that is valid until the destination is reached. Public transport is generally considered cheaper than owning a vehicle thanks to the cost which is distributed across a larger number of people who use the service. Using more of the assets' available capacity leads to a more economical solution for all involved parties despite the typical cost of a bus being greater than a standard family car.

This model can also be applicable to researchers, corporations and educational institutions buying costly equipment without using them at full capacity. Bringing organizations together to share the investment in equipment can reduce cost. However, what prevents more institutions from doing so is the requirement of direct access to physical hardware. An example of this is a biology laboratory which needs to be physically accessible to the researcher. The paper proposes a solution to the problem by implementing a human machine interface which gives users the required control over their equipment whilst being in a remote location.

Just as biologists, robotics reserchers/technicians need to be present on site to utilize their equipment. In the field of robotics, researchers are often bound to be in the same physical location as the hardware. The issue here lies in the accessibility. Too often it happens that additional rework in the physical setup is required before meeting the expected results. Lately, we have seen a rise in remote operations with robotic systems. Whether it concerns educational[1], remote consultancy/debugging[2] or machine learning[3] use-cases, it is prevalent that the human-robot interaction will be increasingly remote.

Remote operations —especially when lead through pro-gramming is employed— are almost always accompanied by delays and inefficiency. Robot simulation software has provided a solution to some extend but the abstraction/assumptions they present hide some of the complexity which is critical to real life applications such as training machine learning systems or achieving complex and highly accurate trajectories.

---
[1]https://www.udacity.com/robot-learning-lab
[2]https://robohub.org/when-you-need-someone-from-canada-to-calibrate-your-robot-in-new-zealand-2/
[3]https://ai.google/research/teams/brain/robotics

The goal of this paper is to propose a setup which allows remote robotic operations to be performed intuitively and without major delaysma. The test case includes an experienced user teaching a movement to a remote robotic arm (similarly to grasping an object) using Dynamic Movement Primitives (DMP). We designed and tested a Human Machine Interface similar to the one shown in Figure 1. The rest of the paper is organized as follows:

- First, we present the related work and describe how our work fits within the literature,
- Next, the system is described and,
- the experimental setup is detailed.
- Finally, we present our results and draw conclusions and plans for future work.

## II. RELATED WORK

Motion planning is required in almost all applications of robotics, spanning from industrial robotic manipulation (e.g. for industrial kitting [1]) to mobile space rovers [2] where there is a need of navigating around obstacles. Other applications include the maximum exploration of environments [3] as well as traversing unstructured environments such as post-disaster scenery [4].

Planning was initially tackled employing graph operations such as Voronoi diagrams [5]. Cell-based approaches were implemented to minimize the computational cost of operations. $A^*$ is based on a fully connected graph [6], while $D^*$ operates on dynamically constructed graphs [7]. Optimal planning concerns the identification not only a viable solutions but, rather, the shortest (or optimal in any other way) one. A review of sampling-based motion planning approaches can be found in [8]. Probabilistic optimal approaches such as the Rapidly-exploring Random Trees (RRT) [9], has gained attraction as it offers probabilistically complete solution with a relatively lightweight operation. Open source implementations of the aforementioned approaches can be found in the Open Motion Planning Library[4] (OMPL).

While optimal planning provides solutions which are efficient, some critical attributes are still missing. Specifically, the solutions are not human-like and thus introduce uncertainties which are not welcome in human-robot collaboration setups. In these cases, learning from demonstration is useful [10], [11]. Moreover, in the era of Industry 4.0 the production lines have become highly adaptive and non-static. Therefore, the motion planning of robot arms have to be adaptive, as well. In these cases, while optimal solutions can enable the robots to reach their desired goals, they tend to produce non-repeatable trajectories.

Dynamic Movement Primitives (DMPs) offer a solution to this problem by allowing adaptability to new goals [12], [13]. DMPs comprise a training phase, when an original trajectory —eg. provided by an expert user using lead-through programming— is employed. Afterwards, when a new goal is received the DMP can provide an adapted trajectory which is similar to the learned one. This is archived by representing the movements as a set of differential equations, which allow for perturbations and parametrization of the start and end goal of the trajectories.

Machine learning has been gaining a lot of attention in the context of arm control. The work in [14] presents a learning approach to the forward dynamics of industrial manipulators, while the more challenging problem on learning inverse dynamics of computed-torque compliant manipulators —such as the Baxter Robot— is presented in [15]. Usually, during remote operations, visual information is utilized to provide feedback. Transmitting video feeds, depending on resolution and quality, can hinder operations. One way to bypass this problem is to employ post processing super-resolution techniques [16] so that the bandwidth requirements are limited.

Computer vision can be computationally expensive and, thus, introduces delays. In cases where resources are scarce and timing is critical, such as space exploration, researchers employ parallelization of computations as in [17]. Computer vision is also employed when there is a need to localize an object —e.g. humans in a scene [18] or landmarks in different views [19]— localize the robot itself with respect to the world around it [20]. Other localization approaches include fixed beacons (anchor points) which provide reference [21].

Computer vision has also be used to capture the motion of humans so that it can be digitized for usage in alternate context —e.g motion capture for animation cinematography [22], [23]. Researchers have been able to detect human hands using only RGB camera images [24] as well as track the hand motions over time [25]. The employment of deep learning [26], [27] has allowed the community to provide extremely accurate results, therefore enabling the employment of human hand actions as a Human-Machine Interface (HMI) [28]. We refer the reader to the paper [29] for an overview of current state of the art in hand detection.

In this work we aim at performing a combination of lead-through programming and learning from demonstration, but on a remote operations setup. The aforementioned related works are vital to our presented system. We employ computer vision for detecting the full 3D pose of human hands, which is used as an input to the HMI. We utilize the OMPL to perform path planning for small trajectories. We employ computer vision for the detection of possible target object for the robot arm. We use DMPs to teach human-like trajectories to the remote robot.

---
[4]http://ompl.kavrakilab.org

## III. SYSTEM DESCRIPTION

The topology of our system (Fig. 1) is spread over two locations, which we'll refer to, hereafter, as Location #1 (*L1*) and Location #2 (*L2*). *L1* and *L2* are connected via a Virtual Private Network (VPN) and communicate using TCP/IP. The purpose of the system is to have a person in location one who can control a robot that is in location two. *L1* includes all the components which are required to allow the user to interact with the system. *L2* includes the robot which is "tele-operated" by the user.

The user provides input to the system by gesturing. The hand of the user —located in *L1*— dictates the location of the robot's end effector. The system identifies the 3D pose of human's hand using the Manomotion's[5] hand tracking solution. The hand tracking software can be executed on low cost devices —such as mobile phones— as it requires only an RGB camera sensor. Manomotion's Software Development Kit (SDK) employs Unity [30] and is also capable of creating intuitive Graphical User Interfaces (GUI). The depth estimation is derived by scale, following a short calibration cycle.

The location coordinates of a hand are sent out to a remote computer/robot —located in *L2*. There, the short movement between the robot's current joint state and the "commanded" state is calculated and executed using OMPL. The objects inside the workspace of the robot are detected by the system automatically and their 3D shape is measured and registered. The resulting joint state of the robot, as well as the objects' location and shape, are sent back to the computer in *L1*.

In *L1*, the user is presented with a visualization of the current state of *L2*, specifically employing RViz [31]. The user continues moving his/her hands until the robot in *L2* has reached its end-goal. This real-time operation is possible because the amount of data exchanged between *L1* and *L2* is minimal, specifically less than a few kilobytes.

When the movement of the robot is completed and the user is satisfied by the smoothness of the trajectory, the training phase is complete. The robot in *L2* sends all the joint state trajectory to *L1* where it is used to train a DMP. The object which was reached in the robot's workspace (in *L2*) is associated with the specific DMP model. This training operation can be repeared for all the objects in the scene.

Following the robot's remote training, whenever an object is detected in the workspace of the robot, the DMP can automatically provide human-like trajectories —as instructed by the user— to reach it.

## IV. EXPERIMENTAL SETUP AND EVALUATION

To assess the system in terms of performance, we have created an experimental setup, capable of both representing our proposed system and of evaluating it's performance by external means. The system comprises multiple components which will be explained below:

[5]https://www.manomotion.com/

1) Robotics Middleware: The framework, on top of which our experimental setup is built, is the Robot Operating System (ROS) [32].
2) User Perception Device: A Phone's mono-camera captures a video stream. The user places his hand in the field of view of the camera to be further utilized by the hand tracking system. The device also provides the hardware to connect to the local wireless network.
3) Hand 3D pose estimation: The video frames are then processed using computer vision to get the the Cartesian position of the user's hand relative to the phone. A solution for this was provided by ManoMotion and is processed on the phone.
4) System Networking: the phone forwards a vector containing the relative hand position and a frame number to the ROS network.
5) Robot: A UR10 robot is set up in a predefined tool-pose and will receive the vector containing the hand position and will add those values to the predefined tool-pose and execute in a loop.
6) Motion Capture System: A high precision, high speed tracking system used to calibrate ManoMotion's SDK to get accurate hand tracking. The employed motion Capture system is the Vicon. The vicon is also used as groundtruth for the evaluation of the system

The evaluation of the proposed system was achieved by running two experiments. Here, we detail the procedure followed during our experiments (see Figure 3), the system calibration method and the steps followed as part of acquiring data. First experiment measures the hand tracking accuracy while the second one compares the robot's traveled distance when trajectories are computed using DMP method against OMPL method. You can see the accuracy results of the 3D hand tracking algorithm vs the Vicon groundtruth in Figure. 2

The accuracy test of the proposed system was conducted using ManoMotion's SDK build on a mobile phone application. The test aims to analyze the system's ability to provide reliable data input for robot tele-operation and trajectory teaching process. It disregards the possible online corrections made by the operator through the visual feedback on robot trajectory. The analysis assumes that the system is adequately calibrated. The experimental setup included a Vicon system used as reference during both calibration phase and collection of hand location data to be compared with the ManoMotion's SDK. The steps undertaken during calibration and data collection are presented in Table I.

Following the training of DMPs on a Universal Robots **UR10**, using our presented remote HRI system, we planned and executed 15 trajectories to 15 randomly chosen end-effector goals. For comparison we planned and executed the same trajectories —i.e. with the same start position and end goals— with OMPL. In order to enhance the reproducibility of our results we chose to use the default settings of the ROS implementations of OMPL and DMP. We then calculated and plotted the total distance travelled by the end-effector in the two cases, as well as the Euclidean distance between the start positions and end goals (Fig. 4). This evaluation provides a twofold insight. First, we can determine how well the DMPs are trained. The closer the distance of the DMP trajectories to the Euclidian distance the better the DMP training, considering that the intended user training was to move in a straight line towards the goal. Secondly, it can be seen that the disparity between DMP and OMPL is significant. Here, exactly, lies the main benefit of our system, it allows the remote training of efficient robot trajectories, instead of executing the —sometimes— unreliable OMPL planned trajectories.

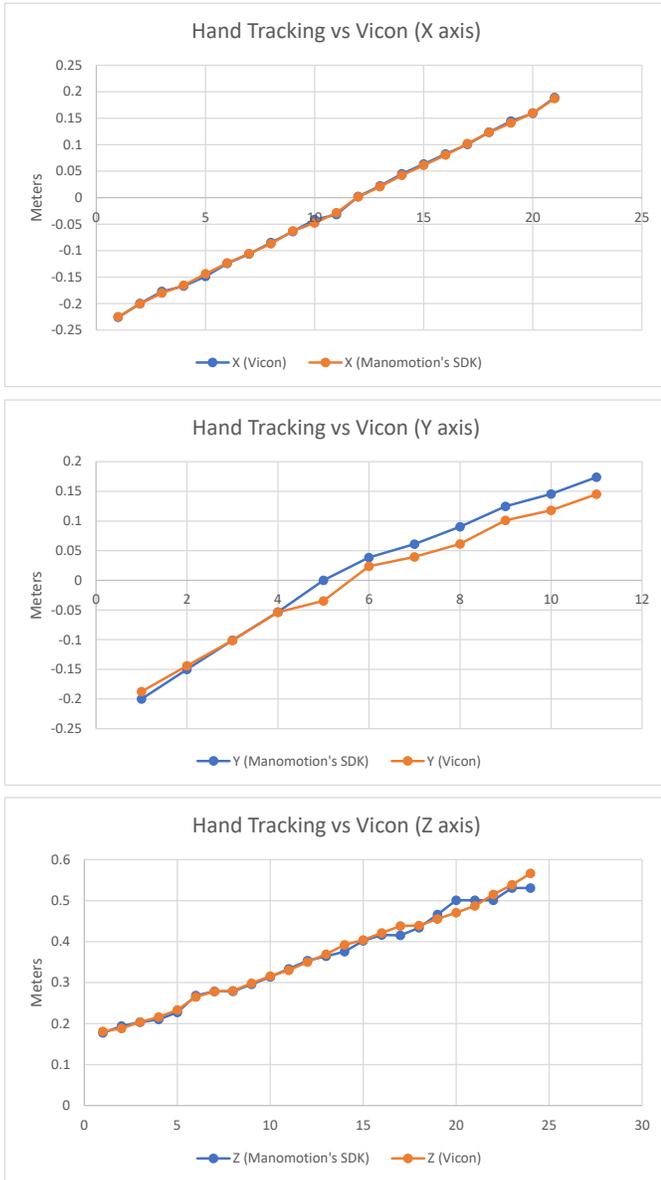

Fig. 2. Evaluation of Hand tracking vs the Vicon groundtruth. The Mean Absolute Deviation was: 0.0085m (X axis), 0.0181m (Y axis), 0.0086m (Z axis)

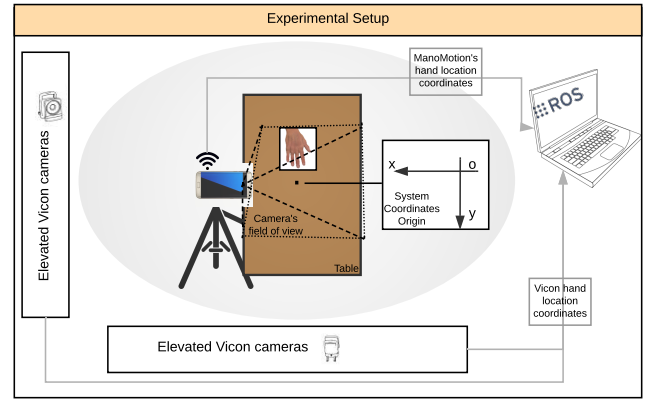

Fig. 3. Experimental setup

| | |
|---|---|
| Experimental Setup Components | 1) Manomotion's SDK build on a Samsung Galaxy S7<br>2) Vicon system and markers<br>3) A user's "hand model" with vicon markers attached to it.<br>4) A tripod with adjustable height<br>5) ROS (Robotic Operating System) |
| Calibration Steps | 1) Alignment of Vicon system coordinates with phone coordinates<br>2) Collection of hand positions data along X, Y, respectively Z axis at constant intervals. The hand location coordinates have been acquired using ManoMotion's SDK, a mobile phone application, and the integrated RGB phone camera The hand location coordinates are an indication of the palm center location<br>3) Simultaneous collection of hand positions using the Vicon system, the reference system<br>4) Regression model computation between ManoMotion's SDK and Vicon data for each of the X, Y and Z axis |
| Accuracy Test Steps | 1) Collection of hand position data along X, Y, respectively Z axis, at constant intervals for analyzing the accuracy on each of the axes individually<br>2) Collection of hand location coordinates, when the hand model was placed at a random location<br>2) Simultaneous collection of hand positions using the Vicon system<br>3) Visual representation of errors<br>4) Computation of MAD (Mean Absolute Deviation) and MAPE (Mean Absolute Percentage Error) |

TABLE I
EXPERIMENTAL SETUP OVERVIEW

## V. CONCLUSION

We presented a novel HRI system capable of efficient and apt training of DMPs on remote robots. The system uses hand gestures –captured by a camera— as an alternative to lead-through programming. The system is inexpensive as it requires low-cost devices such as mobile phones. We evaluated the system using a specifically designed experimental setup. We conclude that the system is promising and should be considered as an option in the upcoming era of remote operations. The investigation of the most optimal hand detection system remains as our future work as well as the trials with multiple users and robot locations.

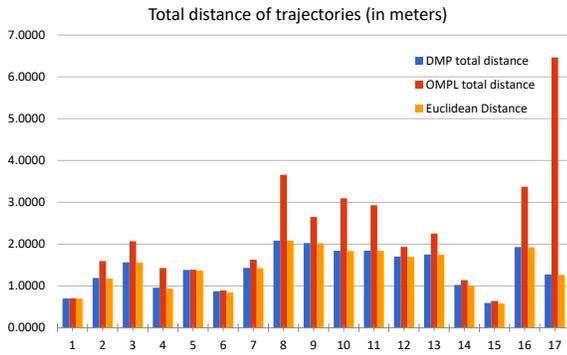

Fig. 4. Evaluation of DMP vs OMPL generated trajectories